\documentclass{article}
\usepackage[dvips]{graphicx}
\usepackage{amsmath,amssymb, bm}
\usepackage{fancybox, enumerate}
\usepackage{my-style}
\usepackage{algpseudocode, algorithm}
\usepackage{multirow}
\newtheorem{theorem}{Theorem}
\allowdisplaybreaks[4]
\newcommand{\myhrulefill}{\leavevmode%
\leaders\hrule depth-2.1pt height 2.5pt\hfill\kern0pt
}
\newcommand{\Mychrulefill}{\leavevmode%
\leaders\hrule depth-2.1pt height 3.5pt\hfill\kern0pt
}

\graphicspath{{figs/}}

\begin{document}
\begin{center}
\Large{\textsc{A Generalization of Spatial Monte Carlo Integration}}\\
\end{center}
\begin{center}
Muneki Yasuda and Kei Uchizawa
\\
Graduate School of Science and Engineering, Yamagata University, Japan
\end{center}

\begin{abstract}
Spatial Monte Carlo integration (SMCI) is an extension of standard Monte Carlo integration 
and can approximate expectations on Markov random fields with high accuracy. 
SMCI was applied to pairwise Boltzmann machine (PBM) learning, with superior results to those from some existing methods. 
The approximation level of SMCI can be changed, 
and it was proved that a higher-order approximation of SMCI is statistically more accurate than a lower-order approximation.
However, SMCI as proposed in the previous studies suffers from a limitation that prevents the application of a higher-order method to dense systems.  

This study makes two different contributions as follows. 
A generalization of SMCI (called generalized SMCI (GSMCI)) is proposed, which allows relaxation of the above-mentioned limitation; 
moreover, a statistical accuracy bound of GSMCI is proved. 
This is the first contribution of this study. 
A new PBM learning method based on SMCI is proposed, which is obtained by combining SMCI and the persistent contrastive divergence. 
The proposed learning method greatly improves the accuracy of learning. 
This is the second contribution of this study. 
\end{abstract}

\section{Introduction}
\label{sec:Intro}

A pairwise Boltzmann machine (PBM)~\cite{Ackley_etal1985} and its variants, 
such as a higher-order Boltzmann machine~\cite{HBM1986}, restricted Boltzmann machine (RBM)~\cite{CD2002,RBM1986}, and deep Boltzmann machine (DBM)~\cite{DBM2009}, 
are one of the most fundamental and important models in the field of probabilistic machine learning. 
Except for some special cases, the inference and learning of PBMs are computationally difficult because they include a multiple summation (or integration) over its all variables. 
Therefore, the development of approximations for them has attracted attention in the field. 
For PBM learning, various methods were proposed, such as, 
mean-field learning methods (e.g., the mean-field approximation~\cite{Kappen1998},
the Bethe approximation (or loopy belief propagation)~\cite{Yasuda2006,Yasuda&Tanaka2009,Federico2012,LBP2012,Cyril2013}, 
the Plefka expansion~\cite{TTanaka1998,Monasson2009}),   
maximum pseudo-likelihood estimation (MPLE)~\cite{PL1975,Hyvarinen2006}, 
contrastive divergence~\cite{CD2002}, 
ratio matching (RM)~\cite{RM2007}, and minimum probability flow (MPF)~\cite{MPF2011}. 

Evaluating the expectations of the variables in PBMs is critical for PBM learning. 
This evaluation is generally NP-hard owing to the multiple summation. 
A Monte Carlo integration (MCI) method is the simplest way to approximate the expectations, 
in which they are approximated by the sample average over the sample points obtained by using a sampling method (e.g., Gibbs sampling) on the PBM.
An effective MCI, called \textit{spatial Monte Carlo integration} (SMCI), was proposed~\cite{Yasuda2015}. 

\begin{figure}[tb]
\begin{center}
\includegraphics[height=3.5cm]{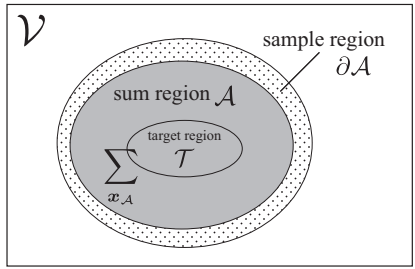}
\end{center}
\caption{Illustration of the target, sum, and sample regions of SMCI.}
\label{fig:region_sheme_SMCI}
\end{figure}

Here, the basic concept of SMCI is informally explained. 
Imagine a PBM, $P(\bm{x})$, defined on a undirected (connected) graph $G(\mcal{V}, \mathbb{F}_2)$ with $n$ vertices, 
where $\bm{x}$ is the set of $n$ variables, $\mcal{V}$ is the set of indices of the vertices, and $\mathbb{F}_2$ is the set of undirected edges. 
For simplicity, the variables are all $\{-1,+1\}$-binary. 
Consider the expectation of $x_i$, i.e., $\ave{x_i} = \sum_{\bm{x}}x_i P(\bm{x})$. 
The exact evaluation of this expectation costs $O(2^n)$. 
In SMCI, this expectation is approximated as follows. 
Suppose that the sample set, $\mathbb{S}$, consists of $M$ sample points generated by a sampling method. 
Take a connected region (or subgraph), $\mcal{A}$, covering vertex $i$, namely, $\{i\} \subseteq   \mcal{A} \subseteq \mcal{V}$. 
For the region $\mcal{A}$, the target expectation is approximated as
\begin{align}
\ave{x_i} \approx \frac{1}{M}\sum_{\ell= 1}^M \sum_{\bm{x}_{\mcal{A}}}x_i P \big(\bm{x}_{\mcal{A}} \mid \bm{x}_{\partial \mcal{A}} = (\text{$\ell$th sample point}) \big),
\label{eqn:concept_SMCI}
\end{align}
where $\bm{x}_{\mcal{A}}$ is the set of variables in $\mcal{A}$, $\bm{x}_{\partial \mcal{A}}$ is the set of variables in the neighborhood of $\mcal{A}$, 
and $P(\bm{x}_{\mcal{A}} \mid \bm{x}_{\partial \mcal{A}} )$ is the conditional distribution of the PBM. 
In SMCI, the target index, $i$, is referred to as the \textit{target region}; 
$\mcal{A}$ and $\partial \mcal{A}$ are referred to as the \textit{sum region} and the \textit{sample region}, respectively. 
The computational cost of equation (\ref{eqn:concept_SMCI}) is $O(M2^{|\mcal{A}|})$, which can be evaluated as long as the size of $\mcal{A}$ is not large. 
Figure \ref{fig:region_sheme_SMCI} illustrates the scheme of the regions of SMCI. 
The formal formulation of equation (\ref{eqn:concept_SMCI}) will be presented in section \ref{sec:GSMCI}.

The primary concern of SMCI is how to determine the sum region. 
The original SMCI~\cite{Yasuda2015} determined the sum region 
so as to cover up to $(k-1)$th nearest neighbors of a specific target region $\mcal{T}$.
This was referred to as the $k$th-order SMCI ($k$-SMCI) method. 
For the approximation accuracy of the $k$-SMCI method, two important statements were proved~\cite{Yasuda2015}:
the $k$th-order SMCI is statistically more accurate than both (i) the standard MCI method for any $k \geq 1$ 
and (ii) the $(k-1)$-SMCI method. 
These statements guarantee that a more accurate approximation can be obtained by increasing the value of $k$ (i.e., the level of approximation of SMCI).
The 1-SMCI method (the simplest $k$-SMCI method) can be applied to any graph as long as the size of the target region is not large, 
because the sum region is identified to the target region in this simplest case. 
However, the 2-SMCI method is not always usable; the sum region can include $O(n)$ variables in a dense graph. 
This is a major drawback of the $k$-SMCI method.

In the original SMCI~\cite{Yasuda2015}, as the level of approximation increases, 
the sum region is systematically expanded according to the neighboring relationship among vertices; 
and this causes the problem mentioned above. 
A more flexible setting of the sum region is desired; for example,  
the sum region covers a part of the first-nearest neighbors of the target region, 
which is called the semi-second-order SMCI (s2-SMCI) method in this paper (cf. section \ref{sec:s2-SMCI}). 
However, accuracy bounds for such sum regions were not clarified.
Intuitively, a larger sum region can be more accurate. 
More concretely, for a specific target region $\mcal{T}$, 
suppose that there are two different regions such that $\mcal{T}\subseteq \mcal{U}_1 \subseteq \mcal{U}_2$; 
in this case, SMCI using $\mcal{U}_2$ as the sum region is more accurate. 
This intuition is in fact true (cf. Theorem \ref{theo:GSMCI}). 
This is the first contribution of this study. 
This type of SMCI is referred to as the generalized SMCI (GSMCI) method in this paper.
From this fact, one can adaptively choose the sum region according to the structure of graph with a clear bound on the approximation accuracy.
In section \ref{sec:SMCI}, the $k$-SMCI method proposed in the previous study is briefly explained.  
The GSMCI method is introduced in section \ref{sec:GSMCI}. 
In these sections, the $k$-SMCI and GSMCI methods are formulated on a higher-order Markov random field (HMRF), 
which is a generalized Markov random field and includes the PBM as a special case. 
The application of the SMCI methods (the $k$-SMCI and GSMCI methods) to the PBM and their numerical validation are presented in section \ref{sec:application_PBM}.

The second contribution of this study is on PBM learning. 
The $k$-SMCI method was applied to PBM learning~\cite{Yasuda2015,SMCI2018} and to another learning problem~\cite{Yasuda2019}. 
In PBM learning, the method based on the 1-SMCI method was superior to the other known learning methods (MPLE, RM, and MPF)~\cite{SMCI2018}. 
In the original learning method~\cite{Yasuda2015,SMCI2018}, the variables in the sample region were fixed by the given training set, 
leading to an useful deterministic algorithm. 
However, the accuracy of the learning cannot be improved without increasing the level of approximation of SMCI (i.e., increasing the value of $k$) in this method. 
In this paper, a new learning method is proposed by combining SMCI with persistent contrastive divergence (PCD)~\cite{PCD2008}. 
This proposed method allows the accuracy of the learning to be improved without increasing the level of approximation of SMCI.
The proposed learning method (with its pseudocode) and its numerical validations are described in section \ref{sec:PBM_learning}.

\section{Spatial Monte Carlo Integration}
\label{sec:SMCI}

In this section, the original SMCI, i.e., the $k$-SMCI method, is explained.

\subsection{Higher-order Markov random field}
\label{sec:HOMRF}

Consider a higher-order MRF (HMRF) consisting of $n$ random variables, $\bm{x} := \{ x_i \in \mcal{X}_i \mid i \in \mcal{V}\}$, 
where $\mcal{X}_i$ is the sample space of $x_i$ and $\mcal{V}:= \{1,2,\ldots, n\}$ is the set of indices of the variables, 
which is defined by
\begin{align}
P(\bm{x}):=\frac{1}{Z} \exp\Big( \sum_{\mcal{C} \in \mathbb{F}} \phi_{\mcal{C}} (\bm{x}_{\mcal{C}})\Big),
\label{eqn:HMRF}
\end{align}
where $\mcal{C}$ denotes a clique, i.e., $\mcal{C}\subseteq \mcal{V}$, and $\mathbb{F}$ denotes a family of cliques. 
A HMRF is regarded as a probabilistic graphical model on an undirected hypergraph, 
in which $\mcal{V}$ is regarded as the set of vertices, and $\mathbb{F}$ is regarded as the set of hyperedges in the hypergraph.
The function $\phi_{\mcal{C}}(\bm{x}_{\mcal{C}})$ denotes a potential function on ${\mcal{C}}$, where $\bm{x}_{\mcal{C}} := \{x_i \mid i \in \mcal{C} \subseteq \mcal{V}\}$.
In equation (\ref{eqn:HMRF}), $Z$ denotes the partition function defined by
\begin{align*}
Z: = \sum_{\bm{x}} \exp\Big( \sum_{\mcal{C} \in \mathbb{F}} \phi_{\mcal{C}} (\bm{x}_{\mcal{C}})\Big),
\end{align*} 
where $\sum_{\bm{x}} := \sum_{x_1 \in \mcal{X}_1}\sum_{x_2 \in \mcal{X}_2}\cdots \sum_{x_n \in \mcal{X}_n}= \prod_{i \in \mcal{V}}\sum_{x_i \in \mcal{X}_i}$ 
denotes the summation over all the possible realizations of $\bm{x}$. 
It should be noted that when $\mcal{X}_i$ is a continuous space, the corresponding sum $\sum_{x_i \in \mcal{X}_i}$ is replaced with the integration $\int_{\mcal{X}_i} dx_i$.
When $\phi_{\mcal{C}}(\bm{x}_{\mcal{C}}) = w_{\mcal{C}} \prod_{i \in \mcal{C}}x_i$, the HMRF is identical to a generalization of a higher-order Boltzmann machine~\cite{HBM1986}. 
In particular, 
when $\mathbb{F} = \mathbb{F}_1 \cup \mathbb{F}_2$, where $\mathbb{F}_1:= \{\{i\} \mid i \in \mcal{V}\}$ 
and $\mathbb{F}_2$ is a family of pairs of indices (i.e., $\mathbb{F}_2:= \{\{i,j\}\}$), and $\phi_{\mcal{C}}(\bm{x}_{\mcal{C}}) = w_{\mcal{C}} \prod_{i \in \mcal{C}}x_i$,  
equation (\ref{eqn:HMRF}) is identical to a familiar PBM~\cite{Ackley_etal1985}:
\begin{align}
P(\bm{x}):=\frac{1}{Z} \exp\Big(\sum_{i \in \mcal{V}}w_i x_i +  \sum_{\{i,j\} \in \mathbb{F}_2} w_{i,j} x_ix_j\Big), 
\label{eqn:pairwiseMRF}
\end{align}
where $w_{i,j}$ is identical to $w_{j,i}$. $w_i$ and $w_{i,j}$ are called the \textit{bias} and \textit{interaction} parameters, respectively.

Here, consider the expectation of a function over specific target variables, $\bm{x}_{\mcal{T}}$, where $\mcal{T}\subseteq \mcal{V}$, on the HMRF, which is 
\begin{align}
\ave{f(\bm{x}_{\mcal{T}})} := \sum_{\bm{x}}f(\bm{x}_{\mcal{T}}) P(\bm{x}) = \sum_{\bm{x}_{\mcal{T}}}f(\bm{x}_{\mcal{T}}) P(\bm{x}_\mcal{T}),
\label{eqn:target_expectation}
\end{align}  
where $P(\bm{x}_{\mcal{T}}) = \sum_{\bm{x} \setminus \bm{x}_{\mcal{T}}}P(\bm{x})$ is the marginal distribution of the HMRF. 
However, in general, the evaluation of this expectation is computationally difficult owing to the multiple summation. 
SMCI, which is an extension of the standard MCI method, was proposed to efficiently approximate the expectation~\cite{Yasuda2015}. 

Hereafter, we assume that the size of $\mcal{T}$ is not large and a sum over $\bm{x}_{\mcal{T}}$ can be numerically evaluated.

\subsection{Adjacency relations in HMRF}

\begin{figure}[tb]
\begin{center}
\includegraphics[height=2.5cm]{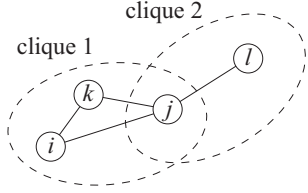}
\end{center}
\caption{Adjacency relations among the indices. For example, $i$ and $j$ are the first-nearest-neighboring indices of each other; 
$i$ and $l$ are the second-nearest-neighboring indices of each other.}
\label{fig:adjacency_relations_HMRF}
\end{figure}

For a detailed explanation of the $k$-SMCI method, an adjacency relation among the indices in $\mcal{V}$ must be defined. 
If a clique $\mcal{C}$ involves both $i$ and $j$, both indices are regarded as the first-nearest-neighboring indices of each other 
(or, in other words, both indices are connected). 
If no cliques involve both $i$ and $j$ and two different cliques involving $i$ and $j$, respectively, overlap, 
both indices are regarded as the second-nearest-neighboring indices of each other (see figure \ref{fig:adjacency_relations_HMRF}). 

Similar to the above, neighboring regions of a target region $\mcal{T} \subseteq \mcal{V}$ are defined as follows.
For the target region $\mcal{T}$, the first-nearest-neighboring region of $\mcal{T}$, $\mcal{N}_1(\mcal{T}) \subseteq \mcal{V}$, is defined as 
\begin{align*}
\mcal{N}_1(\mcal{T}) := \{i \mid i \in \mcal{C} \in \mathbb{F}(\mcal{T}), \> i \not\in \mcal{T}\},
\end{align*}
where $\mathbb{F}(\mcal{A})$ is a subset of $\mathbb{F}$ that is the family of the cliques which overlaps the assigned set $\mcal{A} \subseteq \mcal{V}$ i.e., 
$\mathbb{F}(\mcal{A}) := \{\mcal{C} \mid \mcal{C} \in \mathbb{F},\> \mcal{C} \cap \mcal{A} \neq \emptyset\}$. 
In other words, the indices in $\mcal{N}_1(\mcal{T})$ do not belong to $\mcal{T}$ and simultaneously are the first-nearest neighbors of the indices in $\mcal{T}$. 
The second-nearest-neighboring region of $\mcal{T}$, $\mcal{N}_2(\mcal{T}) \in \mcal{V}$, is defined as
\begin{align*}
\mcal{N}_2(\mcal{T}) := \{i \mid i \in \mcal{C} \in \mathbb{F}(\mcal{N}_1(\mcal{T})), \> i \not\in \mcal{T} \cup \mcal{N}_1(\mcal{T})\}.
\end{align*}
This means that the indices in $\mcal{N}_2(\mcal{T})$ do not belong to $\mcal{T}$ or $\mcal{N}_1(\mcal{T})$ 
and simultaneously are the second-nearest neighbors of the indices in $\mcal{T}$. 
In a similar manner, the $k$th-nearest-neighboring region of $\mcal{T}$, $\mcal{N}_k(\mcal{T}) \subseteq \mcal{V}$, is defined as  
\begin{align}
\mcal{N}_k(\mcal{T}) := \{i \mid i \in \mcal{C} \in \mathbb{F}(\mcal{N}_{k-1}(\mcal{T})), \> i \not\in \mcal{R}_{k-1}(\mcal{T})\},
\label{eqn:R_k(T)}
\end{align}
where $\mcal{R}_{k}(\mcal{T}):= \bigcup_{r=0}^k \mcal{N}_r(\mcal{T})$ and $\mcal{N}_0(\mcal{T}):= \mcal{T}$ i.e., 
$\mcal{R}_{k}(\mcal{T})$ is the region that covers the regions $\mcal{T}, \mcal{N}_1(\mcal{T}), \ldots,\mcal{N}_k(\mcal{T})$. 

\begin{figure}[tb]
\begin{center}
\includegraphics[height=4cm]{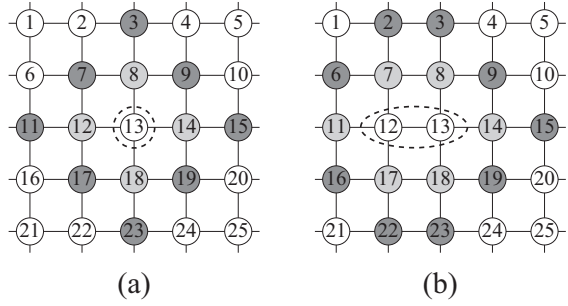}
\end{center}
\caption{Examples of the adjacency relations in a pairwise case:  
(a) when $\mcal{T} = \{13\}$, $\mcal{N}_1(\mcal{T}) =\{ 8,12,14,18\}$ and $\mcal{N}_2(\mcal{T}) =\{ 3,7,9,11,15,17,19,23\}$, 
and (b) when $\mcal{T} =\{12,13\}$, $\mcal{N}_1(\mcal{T}) =\{7,8,11,14,17,18\}$.}
\label{fig:adjacency_relations_MRF}
\end{figure}

The adjacency relations can be easily understand in a pairwise case in which $\mathbb{F} = \mathbb{F}_1 \cup \mathbb{F}_2$. 
In this case, the adjacency relation can be viewed as an undirected graph $G(\mcal{V}, \mathbb{F}_2)$, 
and the adjacency relations represented above are identical to the standard adjacency relations in the graph. 
In figure \ref{fig:adjacency_relations_MRF}, the examples of the adjacency relations in a pairwise case, in which $G(\mcal{V}, \mathbb{F}_2)$ is a square grid graph, are illustrated.

\subsection{$k$th-order SMCI method for HMRF}
\label{sec:k-SMCI}

The spatial Markov property of the HMRF ensures that the conditional distribution 
$P(\bm{x}_{\mcal{R}_{k-1}(\mcal{T})} \mid \bm{x} \setminus \bm{x}_{\mcal{R}_{k-1}(\mcal{T})})$ can be expressed as
\begin{align*}
P(\bm{x}_{\mcal{R}_{k-1}(\mcal{T})} \mid \bm{x} \setminus \bm{x}_{\mcal{R}_{k-1}(\mcal{T})}) 
= P(\bm{x}_{\mcal{R}_{k-1}(\mcal{T})} \mid \bm{x}_{\mcal{N}_{k}(\mcal{T})}),
\end{align*}
where
\begin{align*}
P(\bm{x}_{\mcal{R}_{k-1}(\mcal{T})} \mid \bm{x}_{\mcal{N}_{k}(\mcal{T})}) \propto \exp\Big( \sum_{\mcal{C} \in \mathbb{F}(\mcal{R}_{k-1}(\mcal{T}))} \phi_{\mcal{C}} (\bm{x}_{\mcal{C}})\Big)
\end{align*}
is also an HMRF.
Therefore, the expectation in equation (\ref{eqn:target_expectation}) can be expressed as
\begin{align}
\ave{f(\bm{x}_{\mcal{T}})} = \sum_{\bm{x}_{\mcal{R}_k(\mcal{T})}}f(\bm{x}_{\mcal{T}}) P(\bm{x}_{\mcal{R}_{k-1}(\mcal{T})} \mid \bm{x}_{\mcal{N}_{k}(\mcal{T})}) P(\bm{x}_{\mcal{N}_{k}(\mcal{T})}),
\label{eqn:SMCI_origin}
\end{align}
where $P(\bm{x}_{\mcal{N}_{k}(\mcal{T})})$ is the marginal distribution of $P(\bm{x})$. 

Suppose that $M$ i.i.d. sample points are drawn from $P(\bm{x})$: $\mathbb{S}:=\{ \mbf{s}^{(\ell)} \mid \ell = 1,2,\ldots, M\}$, 
where $\mbf{s}^{(\ell)}:= \{\mrm{s}_i^{(\ell)} \in \mcal{X}_i \mid i \in \mcal{V}\}$ is the $\ell$th sample point. 
In the $k$-SMCI method~\cite{Yasuda2015}, the expectation in equation (\ref{eqn:SMCI_origin}) is approximated by 
\begin{align}
m_{\mcal{T}}^{(k)}(\mathbb{S})
:= \frac{1}{M}\sum_{\ell=1}^M \sum_{\bm{x}_{\mcal{R}_{k-1}(\mcal{T})}}f(\bm{x}_{\mcal{T}}) P\big(\bm{x}_{\mcal{R}_{k-1}(\mcal{T})} \mid \mbf{s}_{\mcal{N}_{k}(\mcal{T})}^{(\ell)} \big),
\label{eqn:k-SMCI}
\end{align}
where $\mbf{s}_{\mcal{A}}^{(\ell)} := \{\mrm{s}_{i}^{(\ell)} \mid i \in \mcal{A}\}$. 
In this approximation, $\mcal{R}_{k-1}(\mcal{T})$ is regarded as the sum region, and $\mcal{N}_{k}(\mcal{T})$ is regarded as the sample region.
Equation (\ref{eqn:k-SMCI}) is obtained by replacing the marginal distribution $P(\bm{x}_{\mcal{N}_{k}(\mcal{T})})$ in equation (\ref{eqn:SMCI_origin}) 
with the corresponding empirical distribution of the given sample points, which is defined by
\begin{align*}
Q_{\mathbb{S}}(\bm{x}_{\mcal{N}_{k}(\mcal{T})}):= \frac{1}{M}\sum_{\ell = 1}^M \delta\big(\bm{x}_{\mcal{N}_{k}(\mcal{T})}, \mbf{s}_{\mcal{N}_{k}(\mcal{T})}^{(\ell)}\big),
\end{align*}
where $\delta$ is the Kronecker (or Dirac) delta function: 
\begin{align*}
\ave{f(\bm{x}_{\mcal{T}})} \approx \sum_{\bm{x}_{\mcal{R}_k(\mcal{T})}}f(\bm{x}_{\mcal{T}}) P(\bm{x}_{\mcal{R}_{k-1}(\mcal{T})} \mid \bm{x}_{\mcal{N}_{k}(\mcal{T})}) Q_{\mathbb{S}}(\bm{x}_{\mcal{N}_{k}(\mcal{T})}) = m_{\mcal{T}}^{(k)}(\mathbb{S}).
\end{align*}
The $k$-SMCI method is usable when the sum over $\bm{x}_{\mcal{R}_{k-1}(\mcal{T})}$ is computable.

\subsection{Asymptotic analysis of $k$-SMCI method}
\label{sec:k-SMCI_asymptotic}

Because the sample points $\mbf{s}^{(\ell)}$ are i.i.d. random variables, 
\begin{align*}
\rho_{\mcal{T}}^{(k)}(\mbf{s}_{\mcal{N}_{k}(\mcal{T})}^{(\ell)}):=\sum_{\bm{x}_{\mcal{R}_{k-1}(\mcal{T})}}f(\bm{x}_{\mcal{T}}) P\big(\bm{x}_{\mcal{R}_{k-1}(\mcal{T})} \mid \mbf{s}_{\mcal{N}_{k}(\mcal{T})}^{(\ell)} \big)
\end{align*}
are also regarded as i.i.d. random variables. 
Therefore, from the result of the central limit theorem, 
the distribution of $m_{\mcal{T}}^{(k)}(\mathbb{S}) = M^{-1} \sum_{\ell = 1}^M \rho_{\mcal{T}}^{(k)}(\mbf{s}_{\mcal{N}_{k}(\mcal{T})}^{(\ell)})$ is asymptotically close to the Gaussian 
with mean 
\begin{align*}
\mu_{\mcal{T}}^{(k)}:= \Big(\prod_{\ell = 1}^M \sum_{\mbf{s}^{(\ell)}}P(\mbf{s}^{(\ell)})\Big) m_{\mcal{T}}^{(k)}(\mathbb{S})
\end{align*}
and variance 
\begin{align*}
v_{\mcal{T}}^{(k)} := \Big(\prod_{\ell = 1}^M \sum_{\mbf{s}^{(\ell)}}P(\mbf{s}^{(\ell)})\Big) m_{\mcal{T}}^{(k)}(\mathbb{S})^2  - \big(\mu_{\mcal{T}}^{(k)}\big)^2
=O(M^{-1}),
\end{align*}
for a sufficiently large $M$, where $\sum_{\mbf{s}^{(\ell)}} := \prod_{i \in \mcal{V}}\sum_{\mrm{s}_i^{(\ell)} \in \mcal{X}_i}$ 
and $P(\mbf{s}^{(\ell)})$ is the HMRF in equation (\ref{eqn:HMRF}).
The asymptotic mean is equivalent to the exact expectation i.e., $\mu_{\mcal{T}}^{(k)}=\ave{f(\bm{x}_{\mcal{T}})}$. 
Therefore, $m_{\mcal{T}}^{(k)}(\mathbb{S})$ converges to $\ave{f(\bm{x}_{\mcal{T}})}$ as $M$ approaches infinity, and its variance vanishes at a speed of $O(M^{-1})$.

From the perspective of statistics, a method having a smaller asymptotic variance is preferable.
For the asymptotic variance $v_{\mcal{T}}^{(k)}$, the following theorem was obtained~\cite{Yasuda2015}. 
\begin{theorem} 
In the HMRF expressed in equation (\ref{eqn:HMRF}), the inequality relation 
$v_{\mcal{T}}\geq v_{\mcal{T}}^{(1)}\geq v_{\mcal{T}}^{(2)}\geq v_{\mcal{T}}^{(3)}\geq \cdots \geq 0$ always holds 
for any $M$ and for any choice of target region $\mcal{T}$, where $v_{\mcal{T}}$ is the asymptotic variance of the standard MCI method.
\label{theo:k-SMCI}
\end{theorem}
This theorem states that, for a sufficient large $M$, the $k$-SMCI method is statistically more accurate than the standard MCI method for any $k\geq 1$ 
and that a higher-order SMCI method is statistically more accurate than any lower-order method.

\section{Generalization of SMCI}
\label{sec:GSMCI}

In this section, the GSMCI method is introduced.
Here, for a region $\mcal{A}$, satisfying $\mcal{T} \subseteq \mcal{A} \subseteq \mcal{V}$, and for a sample set $\mathbb{S}$, consider an SMCI defined as
\begin{align}
m_{\mcal{T}}(\mcal{A}; \mathbb{S}):=
 \frac{1}{M}\sum_{\ell=1}^M\sum_{\bm{x}_{\mcal{A}}}f(\bm{x}_{\mcal{T}}) P\big(\bm{x}_{\mcal{A}} \mid \mbf{s}_{\partial \mcal{A}}^{(\ell)} \big),
\label{eqn:GSMCI}
\end{align}
where $\partial \mcal{A}$ denotes the first-nearest-neighboring region of region $\mcal{A}$: 
$\partial \mcal{A}:=\{i \mid i \in \mcal{C} \in \mathbb{F}(\mcal{A}),\> i \not\in \mcal{A}\}$. 
The conditional distribution in equation (\ref{eqn:GSMCI}) is expressed as
\begin{align}
P\big(\bm{x}_{\mcal{A}} \mid \bm{x}_{\partial \mcal{A}} \big)
\propto \exp\Big( \sum_{\mcal{C} \in \mathbb{F}(\mcal{A})} \phi_{\mcal{C}} (\bm{x}_{\mcal{C}})\Big).
\label{eqn:conditional-dist_GSMCI}
\end{align}
When $\mcal{A} = \mcal{R}_{k-1}(\mcal{T})$, $\partial \mcal{A}$ is identical to $\mcal{N}_k(\mcal{T})$; 
therefore, equation (\ref{eqn:GSMCI}) is equivalent to equation (\ref{eqn:k-SMCI}) in this case. 
Therefore, equation (\ref{eqn:GSMCI}) is regarded as a generalization of the $k$-SMCI method. 
By analogy with the $k$-SMCI method, it is expected that the approximation accuracy of equation (\ref{eqn:GSMCI}) will increase as the sum region $\mcal{A}$ becomes larger. 
In fact, the following argument justifies this expectation.

Consider two sum regions $\mcal{U}_1$ and $\mcal{U}_2$ that satisfy $\mcal{T} \subseteq \mcal{U}_1 \subseteq \mcal{U}_2$. 
For the two sum regions, based on equation (\ref{eqn:GSMCI}), 
the two approximations $m_{\mcal{T}}(\mcal{U}_1; \mathbb{S})$ and $m_{\mcal{T}}(\mcal{U}_2; \mathbb{S})$ 
are considered for the purpose of approximating $\ave{f(\bm{x}_{\mcal{T}})}$.
With a similar argument to that obtained in section \ref{sec:k-SMCI_asymptotic}, 
the asymptotic properties of both approximations are analyzed as follows. 
For a sufficiently large $M$, the distributions of both $m_{\mcal{T}}(\mcal{U}_1; \mathbb{S})$ and $m_{\mcal{T}}(\mcal{U}_2; \mathbb{S})$ 
are asymptotically close to the different Gaussians 
with mean $\mu(\mcal{U}_1)$ and variance $v(\mcal{U}_1)$ and with mean $\mu(\mcal{U}_2)$ and variance $v(\mcal{U}_2)$, respectively.
These asymptotic means and variances are defined as
\begin{align}
\mu(\mcal{A})&:= \Big(\prod_{\ell = 1}^M \sum_{\mbf{s}^{(\ell)}}P(\mbf{s}^{(\ell)})\Big) m_{\mcal{T}}(\mcal{A}; \mathbb{S}) = \ave{f(\bm{x}_{\mcal{T}})}, 
\label{eqn:mu-GSMCI}\\
v(\mcal{A}) &:= \Big(\prod_{\ell = 1}^M \sum_{\mbf{s}^{(\ell)}}P(\mbf{s}^{(\ell)})\Big) m_{\mcal{T}}(\mcal{A}; \mathbb{S})^2  - \mu(\mcal{A})^2,
\label{eqn:v-GSMCI}
\end{align}
for a sum region $\mcal{T} \subseteq \mcal{A} \subseteq \mcal{V}$. 
Equation (\ref{eqn:v-GSMCI}) is rewritten as
\begin{align}
v(\mcal{A})= \frac{1}{M}\Big( \sum_{\bm{x}_{\partial \mcal{A}}} \rho_{\mcal{T}}(\mcal{A}; \bm{x}_{\partial \mcal{A}})^2P(\bm{x}_{\partial \mcal{A}})
- \ave{f(\bm{x}_{\mcal{T}})}^2\Big),
\label{eqn:v-GSMCI_trans}
\end{align}
where
\begin{align}
\rho_{\mcal{T}}(\mcal{A}; \bm{x}_{\partial \mcal{A}}):=
\sum_{\bm{x}_{\mcal{A}}}f(\bm{x}_{\mcal{T}}) P(\bm{x}_{\mcal{A}} \mid \bm{x}_{\partial\mcal{A}} ).
\label{eqn:rho-GSMCI}
\end{align}
Here, $P(\bm{x}_{\partial \mcal{A}})$ and $P(\bm{x}_{\mcal{A}} \mid \bm{x}_{\partial\mcal{A}} )$ are the marginal and conditional distributions of the HMRF of equation (\ref{eqn:HMRF}), 
respectively.
Therefore, it is found that the two approximations $m_{\mcal{T}}(\mcal{U}_1; \mathbb{S})$ and $m_{\mcal{T}}(\mcal{U}_2; \mathbb{S})$ 
converge to the true expectation $\ave{f(\bm{x}_{\mcal{T}})}$ as $M$ approaches infinity, and their variances vanish at speeds of $O(M^{-1})$, as in the $k$-SMCI method. 

For the asymptotic variances $v(\mcal{U}_1)$ and $v(\mcal{U}_2)$, the following theorem is obtained.  
\begin{theorem} 
In the HMRF expressed in equation (\ref{eqn:HMRF}), for $\mcal{T} \subseteq \mcal{U}_1 \subseteq \mcal{U}_2$, the inequality relation 
$v(\mcal{U}_1)\geq v(\mcal{U}_2)\geq 0$ always holds 
for any $M$ and for any choice of target region $\mcal{T}$.
\label{theo:GSMCI}
\end{theorem}
The proof of this theorem is described in \ref{app:proof}. 
This states that $m_{\mcal{T}}(\mcal{U}_2; \mathbb{S})$ is statistically more accurate than $m_{\mcal{T}}(\mcal{U}_1; \mathbb{S})$ for a sufficient large $M$. 
It is noteworthy that this theorem includes the statement of theorem \ref{theo:k-SMCI} as its corollary. 

Equation (\ref{eqn:GSMCI}) is referred to as the GSMCI method in this paper. 
Because equation (\ref{eqn:GSMCI}) is identical to the 1-SMCI method when $\mcal{T} = \mcal{A}$, 
the GSMCI method is statistically more accurate than the standard MCI method for any choice of $\mcal{A}$ satisfying $\mcal{T} \subseteq \mcal{A} \subseteq \mcal{V}$. 
From theorem \ref{theo:GSMCI}, any sum region can be freely selected in equation (\ref{eqn:GSMCI}), and  
it is guaranteed that the approximation accuracy of the GSMCI method will monotonically increases as the size of the selected sum region increases. 
The GSMCI method is usable when the sum over $\bm{x}_{\mcal{A}}$ is computable.

\section{Application to PBMs}
\label{sec:application_PBM}

In this section, the SMCI methods (the $k$-SMCI and GSMCI methods) for the PBM, defined on an undirected graph $G(\mcal{V}, \mathbb{F}_2)$, 
are considered, where the sample spaces of the variables are fixed to $\mcal{X}_i = \{-1,+1\}$. 

The evaluations of $\ave{x_i}$ and $\ave{x_i x_j}$ for $\{i,j\} \in \mathbb{F}_2$ are essential for PBM learning.
From equation (\ref{eqn:k-SMCI}), these expectations are approximated by~\cite{Yasuda2015}
\begin{align}
m_{i}^{(1)}(\mathbb{S})&=\frac{1}{M}\sum_{\ell = 1}^M \tanh \gamma_i\big(\mbf{s}_{\mcal{N}_1(i)}^{(\ell)}\big)
\label{eqn:e1_1-SMCI}
\end{align}
and 
\begin{align}
m_{i,j}^{(1)} (\mathbb{S})&=\frac{1}{M}\sum_{\ell = 1}^M \tanh\big[\atanh\big\{ \tanh\gamma_{i : j}\big(\mbf{s}_{\mcal{N}_1(i,j)}^{(\ell)}\big)
\tanh\gamma_{j : i}\big(\mbf{s}_{\mcal{N}_1(i,j)}^{(\ell)}\big)\big\} + w_{i,j}\big],
\label{eqn:e2_1-SMCI}
\end{align}
respectively, based on the 1-SMCI method, where  
\begin{align}
\gamma_i(\mbf{s}_{\mcal{N}_1(i)}^{(\ell)})&:= w_i + \sum_{j \in \mcal{N}_1(i)}w_{i,j}\mrm{s}_j^{(\ell)}, 
\label{eqn:gamma_i}\\
\gamma_{i : j}(\mbf{s}_{\mcal{N}_1(i,j)}^{(\ell)})&:= \gamma_i(\mbf{s}_{\mcal{N}_1(i)}^{(\ell)}) - w_{i,j}\mrm{s}_j^{(\ell)}, 
\label{eqn:gamma_i-j}
\end{align} 
and $\atanh$ is the inverse hyperbolic tangent function. 
The detailed derivations of equations (\ref{eqn:e1_1-SMCI}) and (\ref{eqn:e2_1-SMCI}) are described in \ref{app:Derivation_1SMCI}.
Equations (\ref{eqn:e1_1-SMCI}) and (\ref{eqn:e2_1-SMCI}) are computable for any $G(\mcal{V}, \mathbb{F}_2)$. 
However, the 2-SMCI methods for $\ave{x_i}$ and $\ave{x_i x_j}$ are not always computable;  
for example, when $G(\mcal{V}, \mathbb{F}_2)$ is a dense graph such as $\mcal{R}_1(\mcal{T}) = O(n)$, 
the computational cost of the evaluation of $m_{\mcal{T}}^{(2)}(\mathbb{S})$ is generally $O(2^n)$ 
because all the variables in $\mcal{R}_1(\mcal{T})$ must be summed over in the 2-SMCI method.

\subsection{Semi-second-order SMCI method}
\label{sec:s2-SMCI}

In the GSMCI method in equation (\ref{eqn:GSMCI}), the sum region $\mcal{A}$ can be freely selected. 
The sum region should be as large as possible within the computational limitation. 
For example, if $\mcal{A}$ is a (cactus) tree, equation (\ref{eqn:GSMCI}) can be computed using a generalized belief propagation~\cite{GBP2001}; 
if $\mcal{A}$ is a planar graph, it can be computed using a combinational technique~\cite{PlanarIsing2016}.
The appropriate choice of the sum region depends on the structure of $G(\mcal{V}, \mathbb{F}_2)$.

In the following, a setting of $\mcal{A}$ that is usable in general cases, semi-second-order SMCI (s2-SMCI) method, is proposed. 
Consider a subset $\mcal{I}_1(\mcal{T}) \subseteq \mcal{N}_1(\mcal{T})$ in which there is no connected (or interacted) pair 
i,e, any two different indices in $\mcal{I}_1(\mcal{T})$ belong to different cliques. 
On the PBM, for $\mcal{A} = \mcal{T} \cup \mcal{I}_1(\mcal{T})$, the conditional distribution in equation (\ref{eqn:conditional-dist_GSMCI}) is represented as  
\begin{align}
P\big(\bm{x}_{\mcal{A}} \mid \bm{x}_{\partial \mcal{A}} \big)
&\propto \exp\Big( \sum_{i \in \mcal{A}}\beta_i\big(\bm{x}_{\partial \mcal{A}}\big)x_i 
+\sum_{\{i,j\} \in \{ c \in \mathbb{F}_2 \mid c \subseteq \mcal{T}\}}w_{i,j}x_ix_j \nn
\aleq
+ \sum_{i \in \mcal{T}}\sum_{j \in \mcal{I}_1(\mcal{T})}w_{i,j} x_ix_j\Big).
\label{eqn:conditional-dist_GSMCI_PBM}
\end{align}
where 
\begin{align}
\beta_i\big(\bm{x}_{\partial \mcal{A}}\big)&:=w_i + \sum_{j \in \partial \mcal{A}}w_{i,j} x_j.
\label{eqn:beta_j}
\end{align}
In equations (\ref{eqn:conditional-dist_GSMCI_PBM}) and (\ref{eqn:beta_j}), $w_{i,j}$'s are regarded as zero when $\{i,j\} \not\in \mathbb{F}_2$. 
The second term of the exponent of equation (\ref{eqn:conditional-dist_GSMCI_PBM}) denotes the interactions in the target region. 
Because the variables in $\mcal{I}_1(\mcal{T})$ do not interact, 
they can be analytically marginalized out from equation (\ref{eqn:conditional-dist_GSMCI_PBM}), leading to 
\begin{align}
P\big(\bm{x}_{\mcal{T}} \mid \bm{x}_{\partial \mcal{A}} \big) 
&=\sum_{\bm{x}_{\mcal{I}_1(\mcal{T})}}P\big(\bm{x}_{\mcal{A}} \mid \bm{x}_{\partial \mcal{A}} \big)\nn
&\propto
\exp\Big\{ \sum_{i \in \mcal{T}}\beta_i\big(\bm{x}_{\partial \mcal{A}}\big)x_i 
+\sum_{\{i,j\} \in \{ c \in \mathbb{F}_2 \mid c \subseteq \mcal{T}\}}w_{i,j}x_ix_j \nn
\aleq
+\sum_{j \in \mcal{I}_1(\mcal{T})} \ln z_j\Big(\beta_j\big(\bm{x}_{\partial \mcal{A}}\big)  + \sum_{i \in \mcal{T}} w_{i,j}x_i \Big)\Big\},
\label{eqn:marginal_conditional-dist_GSMCI_PBM}
\end{align}
where $z_i(a) = \sum_{x_i \in \mcal{X}_i} \exp (a x_i) = 2\cosh a$. 
Therefore, using the marginal distribution in equation (\ref{eqn:marginal_conditional-dist_GSMCI_PBM}), 
the GSMCI method in equation (\ref{eqn:GSMCI}) can be reduced to
\begin{align}
m_{\mcal{T}}(\mcal{A}; \mathbb{S})=
\frac{1}{M}\sum_{\ell=1}^M \sum_{\bm{x}_{\mcal{T}}}f(\bm{x}_{\mcal{T}}) P\big(\bm{x}_{\mcal{T}} \mid \mbf{s}_{\partial \mcal{A}}^{(\ell)} \big),\quad
\mcal{A} = \mcal{T} \cup \mcal{I}_1(\mcal{T}).
\label{eqn:GSMCI_s2}
\end{align}
Equation (\ref{eqn:GSMCI_s2}) is the s2-SMCI method proposed in this section. 
Equation (\ref{eqn:GSMCI_s2}) is computable in a dense graph (as long as the sum over $\bm{x}_{\mcal{T}}$ can be evaluated). 
The s2-SMCI method is regarded as an intermediate approximation between the 1- and 2-SMCI methods 
because it is identical to the 1-SMCI method when $\mcal{I}_1(\mcal{T}) = \emptyset$ 
and is identical to the 2-SMCI method when $\mcal{I}_1(\mcal{T}) = \mcal{N}_1(\mcal{T})$; 
therefore, from the result obtained in theorem \ref{theo:GSMCI}, 
the approximation accuracy of the s2-SMCI method should be intermediate between those of the 1- and 2-SMCI methods.   

\begin{figure}[tb]
\begin{center}
\includegraphics[height=3cm]{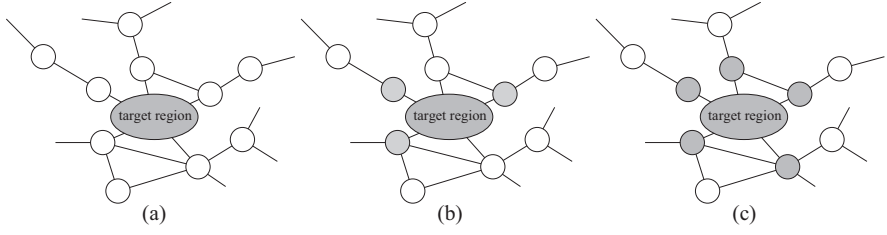}
\end{center}
\caption{Illustration of the sum regions of the 1-, s2-, and 2-SMCI methods (the sum regions are shaded): 
(a) the 1-SMCI method, (b) an example of the s2-SMCI method, and (c) the 2-SMCI method. 
In the s2-SMCI method, another choice of the sum region is possible.}
\label{fig:s2-SMCI}
\end{figure}
 
Using equations (\ref{eqn:marginal_conditional-dist_GSMCI_PBM}) and (\ref{eqn:GSMCI_s2}),  
the s2-SMCI methods for the expectations, $\ave{x_i}$ and $\ave{x_i x_j}$ ($\{i,j\} \in \mathbb{F}_2$), 
are expressed as
\begin{align}
m_{i}(\mcal{A}; \mathbb{S})&=\frac{1}{M}\sum_{\ell=1}^M \tanh \xi_{i} \big(\mbf{s}_{\partial \mcal{A}}^{(\ell)} \big)
\label{eqn:e1_s2-SMCI}
\end{align}
and
\begin{align}
m_{i,j}(\mcal{A}; \mathbb{S})&=\frac{1}{M}\sum_{\ell=1}^M \tanh\big[\atanh \big\{ \tanh \xi_{i : j} \big(\mbf{s}_{\partial \mcal{A}}^{(\ell)} \big)
\tanh\xi_{j : i} \big(\mbf{s}_{\partial \mcal{A}}^{(\ell)} \big)\big\} + \omega_{i,j}\big(\mbf{s}_{\partial \mcal{A}}^{(\ell)} \big)\big],
\label{eqn:e2_s2-SMCI}
\end{align}
respectively, where
\begin{align}
\xi_{i}(\mbf{s}_{\partial \mcal{A}}^{(\ell)})&:=\beta_i\big(\mbf{s}_{\partial \mcal{A}}^{(\ell)}\big) + \sum_{j \in \mcal{I}_1(i)}\atanh \big\{ \tanh \beta_j\big(\mbf{s}_{\partial \mcal{A}}^{(\ell)}\big) \tanh w_{i,j}\big\},
\label{eqn:xi_i}\\
\xi_{i : j}(\mbf{s}_{\partial \mcal{A}}^{(\ell)})&:=\beta_i\big(\mbf{s}_{\partial \mcal{A}}^{(\ell)}\big) 
+ \sum_{k \in \mcal{I}_1(i,j)}\atanh \big\{ \tanh \beta_k\big(\mbf{s}_{\partial \mcal{A}}^{(\ell)}\big) \tanh w_{i,k}\big\}\nn
\aldef
+\frac{1}{4} \sum_{k \in \mcal{I}_1(i,j)}\ln \frac{1 - \tanh^2\{\beta_k(\mbf{s}_{\partial \mcal{A}}^{(\ell)}) + w_{i,k}\} \tanh^2 w_{j,k}}
{1 - \tanh^2\{\beta_k(\mbf{s}_{\partial \mcal{A}}^{(\ell)}) - w_{i,k}\} \tanh^2 w_{j,k}},
\label{eqn:xi_i-j}\\
\omega_{i,j}\big(\mbf{s}_{\partial \mcal{A}}^{(\ell)} \big)&:=w_{i,j}
+ \sum_{k \in \mcal{I}_1(i,j)}\atanh\big\{ \tanh w_{i,k} \tanh w_{j,k}\big\}\nn
\aldef
+\frac{1}{4} \sum_{k \in \mcal{I}_1(i,j)}\ln \frac{1 - \tanh^2(w_{i,k} + w_{j,k}) \tanh^2 \beta_k(\mbf{s}_{\partial \mcal{A}}^{(\ell)})}
{1 - \tanh^2(w_{i,k} - w_{j,k}) \tanh^2 \beta_k(\mbf{s}_{\partial \mcal{A}}^{(\ell)})}.
\label{eqn:pmega_ij}
\end{align}
Equations (\ref{eqn:e1_s2-SMCI}) and (\ref{eqn:e2_s2-SMCI}) are computable in a dense graph. 
In equations (\ref{eqn:xi_i})--(\ref{eqn:pmega_ij}), $w_{i,j}$'s are zero when $\{i,j\} \not\in \mathbb{F}_2$. 
The detailed derivations of equations (\ref{eqn:e1_s2-SMCI}) and (\ref{eqn:e2_s2-SMCI}) are described in \ref{app:Derivation_s2SMCI}.

From the result obtained in theorem \ref{theo:GSMCI}, a larger $\mcal{I}_1(\mcal{T})$ is preferable. 
However, the maximization of the size of $\mcal{I}_1(\mcal{T})$ is known as the \textit{maximum independent set} (MIS) problem, which is an NP-hard optimization problem. 
The well-known greedy algorithm~\cite{MIS1997} for this problem is presented in Algorithm \ref{alg:greedy-MIS}. 
In this algorithm, $\mrm{deg}(i;\mcal{U})$ denotes the degree of $i$ in the subgraph $\mcal{U}$, 
and $\partial_{\mcal{U}}i:=\{j \mid j \in \mcal{U}, \> \{i,j\} \in \mathbb{F}_2\}$ 
denotes the set of the indices that are the first-nearest neighbors of $i$ in the subgraph $\mcal{U}$ i.e., $\mrm{deg}(i;\mcal{U}) = |\partial_{\mcal{U}}i|$. 
\begin{algorithm}[H]
\caption{Greedy algorithm for MIS problem~\cite{MIS1997}}
\label{alg:greedy-MIS}
\begin{algorithmic}[1]
\State \textbf{Input} $\mcal{N}_1(\mcal{T}) \subseteq \mcal{V}$ and $\mathbb{F}_2$
\State $\mcal{I}_1(\mcal{T}) \gets \emptyset$ and $\mcal{U} \gets \mcal{N}_1(\mcal{T})$
\Repeat
\State Choose index $r$ such that $r = \argmin_{j \in \mcal{U}} \mrm{deg}(j;\mcal{U})$
\State $\mcal{I}_1(\mcal{T}) \gets \mcal{I}_1(\mcal{T}) \cup \{r\}$
\State $\mcal{U} \gets \mcal{U} \setminus \big(\partial_{\mcal{U}}r \cup \{r\} \big)$
\Until{$\mcal{U} \neq \emptyset$}
\State \textbf{Output} $\mcal{I}_1(\mcal{T})$.
\end{algorithmic}
\end{algorithm}

In step 4 in Algorithm \ref{alg:greedy-MIS}, one may encounter the case in which  
multiple indices have the same minimum degree. 
They are the equal candidates of $r$. 
When the minimum degree is zero, the selection does not affect the final result 
because all the candidates will be included in $\mcal{I}_1(\mcal{T})$. 
However, when the minimum degree is larger than zero, one of them must be selected, and the selection can affect the final result.
A criterion for the selection is needed.
In this paper, a heuristic for the selection is proposed: 
when multiple indices have the same minimum degree that is larger than zero, 
the index that has the maximum $W_j$ is selected, where $W_j := \sum_{i \in \mcal{T} \cap \partial j}|w_{i,j}|$. 
$W_j$ is regarded as the absolute strength of interaction between index $j$ and the target region. 
This heuristic is based on our usual sense; that is, a pair having stronger interaction is more important. 

\subsection{Experiment}

In this section, the validity of the proposed method is demonstrated using numerical experiments. 
PBMs of $|\mcal{V}| = n= 20$ defined on two types of undirected graphs $G(\mcal{V}, \mathbb{F}_2)$ were used: a $4\times 5$ square grid graph and a random graph with connection probability $p$. 
The bias and interaction parameters, $w_i$ and $w_{i,j}$, in the PBMs were generated from uniform distributions having the intervals $[-0.2,+0.2]$ and $[-0.3,+0.3]$, respectively.
On the PBMs, the approximation accuracies of the SMCI methods (the 1-, s2-, and 2-SMCI methods) were checked. 
The accuracy of the approximation was measured by the mean absolute error (MAE) of the covariances: 
\begin{align*}
\mrm{MAE}=\frac{1}{|\mathbb{F}_2|} \sum_{\{i,j\} \in \mathbb{F}_2}\big| \chi_{i,j}^{\mrm{exact}} - \chi_{i,j}^{\mrm{approx}}\big|,
\end{align*}
where $\chi_{i,j}^{\mrm{exact}}:= \ave{x_ix_j} - \ave{x_i}\ave{x_j}$ is the exact covariance, and $\chi_{i,j}^{\mrm{approx}}$ is its approximation. 
Because the size of $\mcal{V}$ is not large, the exact covariance can be numerically evaluated. 
The sample sets, $\mathbb{S}$, used in the SMCI methods were generated from the PBM by using Gibbs sampling based on a simulated annealing.

\begin{figure}[tb]
\begin{center}
\includegraphics[height=3.7cm]{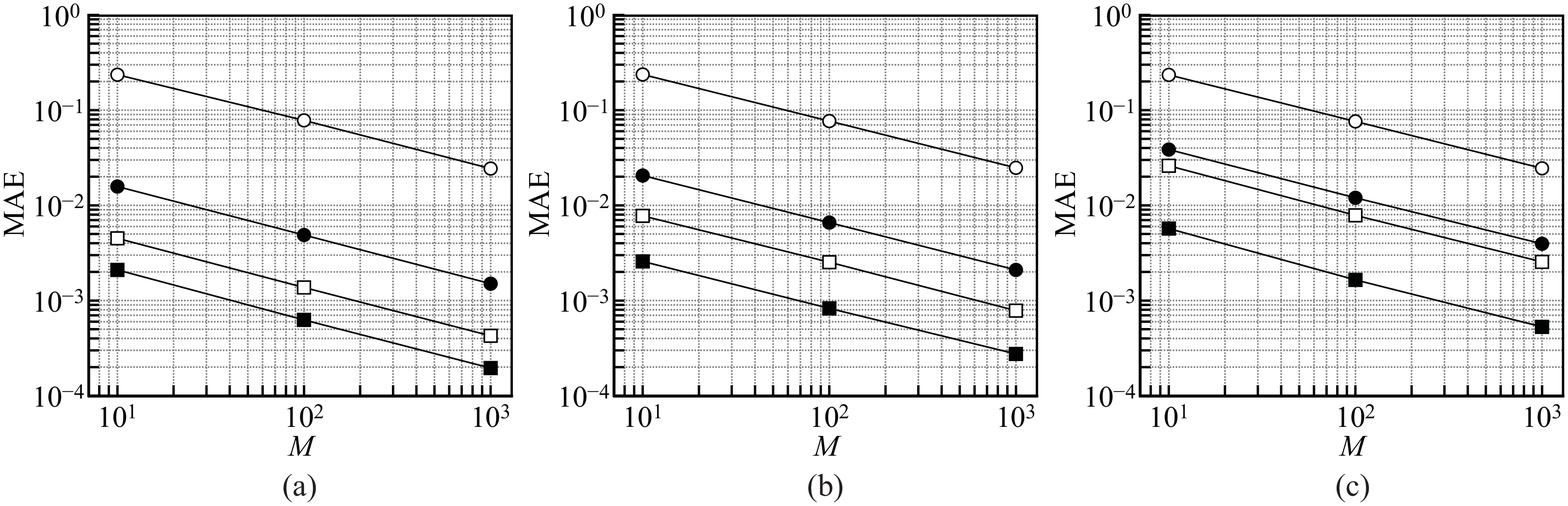}
\end{center}
\caption{MAEs for $M = 10,100,1000$ in (a) a $4\times 5$ square grid graph, (b) a random graph of $p = 0.2$, 
and (b) a random graph of $p = 0.4$. 
The open and filled circles, ``$\circ$'' and ``$\bullet$'', denote the standard MCI and 1-SMCI methods, respectively; 
the open and filled squares, ``{\tiny $\square$}'' and ``{\tiny $\blacksquare$}'', denote the s2- and 2-SMCI methods, respectively. 
The plots are the average values over 200 experiments.}
\label{fig:inf_result}
\end{figure}

The plots in Figure \ref{fig:inf_result} show the results in (a) a $4\times 5$ square grid graph, (b) a random graph of $p = 0.2$, 
and (c) a random graph of $p = 0.4$, for various $M$. 
For the comparison, the results obtained by the standard MCI method are also plotted. 
In the 2-SMCI method, the sum over $\bm{x}_{\mcal{R}_1(\mcal{T})}$ was numerically evaluated.
As expected, the SMCI methods are superior to the standard MCI method, and the accuracy of the s2-SMCI method is intermediate between those of the 1- and 2-SMCI methods. 
The accuracies of the SMCI methods in (c) are worse than those in (a) and (b). 
This indicates that they are more effective in a sparser graph. 
In figures \ref{fig:inf_result}(a)--(c), the accuracy of the standard MCI method of $M = 1000$ is almost the same that of the 1-SMCI method of $M = 10$;  
therefore, the standard MCI method needs an approximately 100 times larger sample set to reach the accuracy of the 1-SMCI method (at least in the settings of the present experiments). 
The MAEs were also evaluated by using annealed importance sampling (AIS)~\cite{Neal2001}. 
The results of AIS were almost the same as those of the standard MCI method. 
In the AIS, the initial distribution was set to an uniform distribution over $\{-1,+1\}^n$, 
and a sequence of the inverse temperature, $0 = \beta_0 < \beta_1 < \cdots < \beta_K = 1$, was set as $\beta_{k+1} = \beta_k + 10^{-4}$.

\section{Application to PBM Learning}
\label{sec:PBM_learning}

In this section, the learning of the PBM defined on $G(\mcal{V}, \mathbb{F}_2)$ is considered. 
The PBM is represented with the explicit dependency on its parameters $P(\bm{x}) = P(\bm{x} \mid \theta)$, 
where $\theta$ denotes the set of the parameters of the PBM, $\theta:= \{w_i ,w_{i,j} \mid i \in \mcal{V},\, \{i,j\} \in \mathbb{F}_2\}$.
Suppose that a training dataset consisting of $N$ training data points, $\mathbb{D}:=\{ \mbf{x}^{(\mu)} \mid \mu = 1,2,\ldots, N\}$, is obtained. 
For the dataset, the log likelihood is defined as  
\begin{align}
L_{\mathbb{D}}(\theta):=\frac{1}{N} \sum_{\mu = 1}^N \ln P\big(\mbf{x}^{(\mu)} \mid \theta \big). 
\label{eqn:log-likelihood}
\end{align} 
The log likelihood is the function with respect to $\theta$. 
PBM learning is performed via maximum likelihood estimation (MLE), 
that is, by maximizing the log likelihood with respect to $\theta$. 
The parameters in $\theta$ are referred as to the learning parameters.
The gradients of the log likelihood with respect to $w_i$ and $w_{i,j}$, for $\mathbb{D}$, are 
\begin{align}
g_i(\mathbb{D},\theta)&:= \frac{\partial L_{\mathbb{D}}(\theta)}{\partial w_i}=\frac{1}{N}\sum_{\mu = 1}^N \mrm{x}_i^{(\mu)} - \ave{x_i}, \\
g_{i,j}(\mathbb{D},\theta)&:=\frac{\partial L_{\mathbb{D}}(\theta)}{\partial w_{i,j}}=\frac{1}{N}\sum_{\mu = 1}^N \mrm{x}_i^{(\mu)}\mrm{x}_j^{(\mu)} - \ave{x_ix_j},
\end{align}
respectively. These gradients have the intractable expectations in their second terms. 

An approximation based on the $k$-SMCI method was proposed~\cite{Yasuda2015}, 
in which the intractable expectations are approximated by the $k$-SMCI method i.e., 
$\ave{x_i} \approx m_i^{(k)}(\mathbb{D})$ and $\ave{x_ix_j} \approx m_{i,j}^{(k)}(\mathbb{D})$. 
In this approximation, the sample set $\mathbb{S}$ is fixed to the dataset. 
This approximation method based on the 1-SMCI method provides better learning results 
than some known learning algorithms~\cite{SMCI2018}: MPLE~\cite{PL1975}, 
RM~\cite{RM2007}, and MPF~\cite{MPF2011}. 

In the previous learning method~\cite{Yasuda2015}, the variables in the sample region were fixed by the given training set, 
leading to an useful deterministic algorithm. 
However, the accuracy of the learning cannot be improved without increasing the level of approximation of SMCI (i.e., increasing the value of $k$) in this method. 
Moreover, the accuracy of the learning was degraded in a model-mismatched case~\cite{SMCI2018}, 
in which the graph structures of the data generating PBM and those of the learning PBM were different (or more precisely,  
the graph of the learning PBM did not have that of the generating PBM as a subgraph). 
This degradation was caused by fixing the sample set to the dataset. 
An SMCI method is obtained by approximating the marginal distribution over the sample region by the sample distribution (cf. section \ref{sec:k-SMCI}).
In a model-mismatched case, the data distribution is no longer a good approximation of the corresponding marginal distribution in the sample region. 

\subsection{Proposed learning algorithm}
\label{sec:proposed_learning}

\begin{figure}[tb]
\begin{center}
\includegraphics[height=5cm]{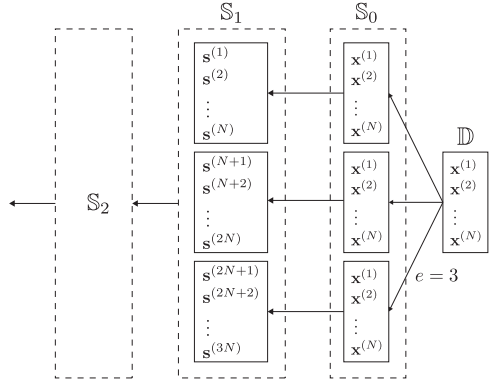}
\end{center}
\caption{Illustration of the update procedure of the sample set when $e = 3$. $\mathbb{S}_0$ is set to the $e$-replicated $\mathbb{D}$. 
Each sample point in $\mathbb{S}_t$ is updated via $\kappa$-steps Gibbs sampling on $P(\bm{x} \mid \theta_{t+1})$; 
therefore, $M (= eN)$ parallel Gibbs sampling is run to update the sample set.}
\label{fig:sample_procedure}
\end{figure}

The present paper proposes a new approximation for PBM learning, which is explained as follows. 
At first, a sample set $\mathbb{S}$ with $M = eN$ is prepared, where $M$ is the size of the sample set, and $e$ is a positive integer called the ``data-extension rate.'' 
The sample set and the learning parameters are initialized as $\mathbb{S} = \mathbb{S}_0$ and $\theta = \theta_0$, respectively.
Using $\mathbb{S}_0$, the learning parameters $\theta_0$ are updated to $\theta_1$ based on a gradient ascent method with the approximate gradients 
$g_i^{\mrm{app}}(\mathbb{S}_0, \theta_0)$ and $g_{i,j}^{\mrm{app}}(\mathbb{S}_0, \theta_0)$ defined by 
\begin{align}
g_i^{\mrm{app}}(\mathbb{S}, \theta)&:= \frac{1}{N}\sum_{\mu = 1}^N \mrm{x}_i^{(\mu)} - m_i(\mcal{A}_i;\mathbb{S}), \\
g_{i,j}^{\mrm{app}}(\mathbb{S}, \theta)&:= \frac{1}{N}\sum_{\mu = 1}^N \mrm{x}_i^{(\mu)}\mrm{x}_j^{(\mu)} - m_{i,j}(\mcal{A}_{i,j};\mathbb{S}),
\end{align}
where $m_i(\mcal{A}_i;\mathbb{S})$ and $m_{i,j}(\mcal{A}_{i,j};\mathbb{S})$ are the approximations of $\ave{x_i}$ and $\ave{x_i x_j}$ on $P(\bm{x} \mid \theta)$, 
respectively, based on the GSMCI method proposed in equation (\ref{eqn:GSMCI}); 
here, $\mcal{A}_i$ and $\mcal{A}_{i,j}$ are the sum regions determined for the corresponding target regions.
After the update of $\theta$, the sample set $\mathbb{S}_0$ is updated to $\mathbb{S}_1$ by using ($M$ parallel) $\kappa$-steps Gibbs sampling on $P(\bm{x} \mid \theta_1)$. 
By using $\mathbb{S}_1$, the learning parameters are again updated to $\theta_2$ using the gradients $g_i^{\mrm{app}}(\mathbb{S}_1, \theta_1)$ and $g_{i,j}^{\mrm{app}}(\mathbb{S}_1, \theta_1)$.
This two-stages updating procedure, i.e., the parameter update and sample set update stages, is repeated during learning: 
$\mathbb{S}_0, \theta_0 \to \theta_1 \to \mathbb{S}_1 \to \theta_2 \to \mathbb{S}_2 \to \cdots$.  
The proposed procedure is inspired by the PCD method~\cite{PCD2008}. 
The initial state of the sample set, i.e., $\mathbb{S}_0$, is set to the $e$-replicated $\mathbb{D}$. 
The update procedure of the sample set is illustrated in figure \ref{fig:sample_procedure}.
The pseudocode of the proposed learning is presented in Algorithm \ref{alg:learning}.

\begin{algorithm}[H]
\caption{The proposed learning algorithm for PBM}
\label{alg:learning}
\begin{algorithmic}[1]
\State \textbf{Input} training dataset $\mathbb{D}$
\State Initialize the learning parameters: $\theta = \theta_0$
\State Initialize the sample set by using the $e$-replicated $\mathbb{D}$: $\mathbb{S} = \mathbb{S}_0$ 
\State Set $t = 0$
\Repeat
\State Update the learning parameters using a gradient ascent method with $g_i^{\mrm{app}}(\mathbb{S}_t, \theta_t)$ and $g_{i,j}^{\mrm{app}}(\mathbb{S}_t, \theta_t)$; e.g., 
\begin{align*}
w_i^{(t+1)} &\gets w_i^{(t)} + \varepsilon g_i^{\mrm{app}}(\mathbb{S}_t, \theta_t),\nn
w_{i,j}^{(t+1)} &\gets w_{i,j}^{(t)} + \varepsilon g_{i,j}^{\mrm{app}}(\mathbb{S}_t, \theta_t),
\end{align*}
where $\varepsilon$ is the learning rate.
\State Update the sample set $\mathbb{S}_{t}$ to $\mathbb{S}_{t+1}$ using ($M$ parallel) $\kappa$-steps Gibbs sampling on $P(\bm{x} \mid \theta_{t+1})$ 
(starting from $\mathbb{S}_{t}$)
\State $t \gets t + 1$
\Until{A certain criterion is satisfied}
\end{algorithmic}
\end{algorithm}

\subsection{Experiment}

In this section, the performance of the proposed learning algorithm described in the previous section is demonstrated. 
In the following experiments, the training datasets of $N = 50$ were generated from a generative PBM (g-PBM) with $n = 20$, 
which has the same form as equation (\ref{eqn:pairwiseMRF}), by using Gibbs sampling. 
A training PBM (t-PBM) having the same size as g-PBM was trained using the generated artificial datasets.
Two cases are considered: 
(i) the model-matched case in which g-PBM is defined on a $4 \times 5$ square grid graph and t-PBM is also defined on the same square grid graph; 
and (ii) the model-mismatched case in which g-PBM is defined on a fully connected graph and t-PBM is defined on a $4 \times 5$ square grid graph. 
Because the size of t-PBM is not very large, MLE can be performed exactly.   
The parameters $w_i$ and $w_{i,j}$ in g-PBM were randomly selected according to uniform distributions having the intervals $[-0.2,+0.2]$ and $[-0.3,+0.3]$, respectively. 
In the following experiments, $\kappa$ was fixed to one, and the learning rate was fixed to $0.02$.

\begin{figure}[tb]
\begin{center}
\includegraphics[height=4.4cm]{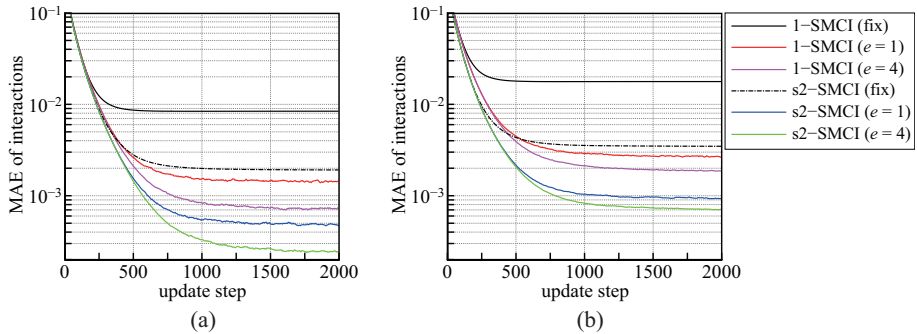}
\end{center}
\caption{
MAEs of interactions in (a) the model-matched case and (b) the model-mismatched case, versus the update step $t$.
The plots are the average values over 200 experiments.}
\label{fig:train_result}
\end{figure}

The accuracy of the learning was measured by the MAE of the interactions: 
\begin{align*}
\mrm{MAE}(t)=\frac{1}{|\mathbb{F}_2|} \sum_{\{i,j\} \in \mathbb{F}_2}\big| w_{i,j}^{(t)} - w_{i,j}^{\mrm{MLE}}\big|,
\end{align*}
where $w_{i,j}^{\mrm{MLE}}$ is the value obtained from the exact MLE, and $w_{i,j}^{(t)}$ is the trained value at step $t$. 
Figures \ref{fig:train_result}(a) and (b) present the MAEs against the update step $t$; 
(a) is the result of the (i) the model-matched case and (b) is that of (ii) the model-mismatched case. 
In these figures, ``1-SMCI'' and ``s2-SMCI'' correspond to the training based on the 1-SMCI and s2-SMCI methods.  
``(fix)'' means that the sample sets were fixed to the dataset ($\mathbb{S} = \mathbb{D}$), i.e., the learning strategy proposed in the previous studies~\cite{Yasuda2015,SMCI2018}; 
the others are the proposed method combined with the PCD-like strategy in which $e$ is the data-extension rate. 
The proposed method greatly improved the accuracy in both cases. 
The accuracies in the model-mismatched case were worse than those in the model-matched case.  
However, the proposed method can reduce the accuracy degradation by increasing the value of $e$.

\section{Summary and Future Works}

In this paper, two different contributions for PBM were presented. 
The first contribution is a generalization of the original SMCI method~\cite{Yasuda2015}, described in section \ref{sec:GSMCI}. 
In the original SMCI method (i.e., the $k$-SMCI method), the setting of the sum region was seriously limited; i.e., 
for a target region $\mcal{T}$, the sum region must cover up to the $(k-1)$th-nearest-neighboring region, $\mcal{R}_{k-1}(\mcal{T})$, of the target region. 
The statistical accuracy bound of the $k$-SMCI method was proved~\cite{Yasuda2015}(cf. Theorem \ref{theo:k-SMCI}). 
However, a higher-order $k$-SMCI method cannot be applied in a dense graph 
because the size of the sum region can be $O(n)$ there. 
This study investigated a more flexible setting of the sum region and provided a statistical accuracy bound of the setting (cf. Theorem \ref{theo:GSMCI}).  
The proposed method (i.e, the GSMCI method) allows a flexible setting of the sum region, such as in the s2-SMCI method discussed in section \ref{sec:s2-SMCI}. 
The statistical accuracy bounds of the $k$-SMCI and GSMCI methods were validated in generalized MRFs.

The second contribution of this study is a new algorithm for PBM learning, described in section \ref{sec:PBM_learning}. 
The proposed learning method greatly improved the accuracies of learning in the model-matched and model-mismatched cases. 
The learning method proposed by the previous study~\cite{Yasuda2015,SMCI2018} is applicable to only fully visible PBMs, 
because the values of the variables in the sample region must be filled by the dataset.
The proposed learning method can be immediately applied to PBM learning with hidden variables, 
such as RBMs (and its variants: Gaussian-Bernoulli RBMs~\cite{GBRBM2011} and Gaussian-Spherical RBMs~\cite{GSRBM2020}) and DBMs, 
because the values of the (hidden) variables in the sample region are filled by the sample points obtained by Gibbs sampling.
Application to RBMs and DBMs is considered as important future works.

SMCI uses a sample set drawn from an MRF; therefore, the approximation accuracy of SMCI depends on the quality of the sampling algorithm. 
In the experiments of this paper, usual Gibbs sampling was used. 
An effective sampling method, based on belief propagation, was proposed~\cite{BP+MCMC2014}, the ideas of which is close to SMCI. 
It is believed that a combination of both SMCI and this sampling method leads to more effective approximations. 
This is also considered as an important future work.

\subsection*{Acknowledgment}
This work was partially supported by JSPS KAKENHI (Grant Numbers 15H03699, 18K11459, and 18H03303), 
JST CREST (Grant Number JPMJCR1402), and the COI Program from the JST (Grant Number JPMJCE1312).  

\appendix

\section{Proof of Theorem \ref{theo:GSMCI}}
\label{app:proof}

From the definition, the asymptotic variance $v(\mcal{A})$ for any region $\mcal{A}$ satisfying $\mcal{T} \subseteq \mcal{A} \subseteq \mcal{V}$ 
is always greater than or equal to zero.
In the following, the difference between the asymptotic variances $v(\mcal{U}_1)$ and $v(\mcal{U}_2)$, 
$e:= M(v(\mcal{U}_1)- v(\mcal{U}_2))$, is considered, where $\mcal{T} \subseteq \mcal{U}_1 \subseteq \mcal{U}_2$.
From equation (\ref{eqn:v-GSMCI_trans}), 
the difference is represented as
\begin{align}
E = \sum_{\bm{x}_{\partial\mcal{U}_1}}\rho_{\mcal{T}}(\mcal{U}_1;\bm{x}_{\partial \mcal{U}_1})^2 P(\bm{x}_{\partial\mcal{U}_1})
-\sum_{\bm{x}_{\partial\mcal{U}_2}}\rho_{\mcal{T}}(\mcal{U}_2;\bm{x}_{\partial \mcal{U}_2})^2 P(\bm{x}_{\partial\mcal{U}_2}).
\label{eqn:diff_rho}
\end{align}
For the evaluation of equation (\ref{eqn:diff_rho}), $\mcal{Y}:=\mcal{U}_2 \setminus \mcal{U}_1$ is defined. 
The relation $\partial \mcal{U}_1 \subseteq \partial \mcal{U}_2 \cup \mcal{Y}$ is satisfied because 
\begin{align}
\partial \mcal{U}_1 &= \{i \mid i \in \mcal{C} \in \mathbb{F}(\mcal{U}_1),\> i \not\in \mcal{U}_1\} \nn
&\subseteq \{i \mid i \in \mcal{C} \in \mathbb{F}(\mcal{U}_2),\> i \not\in \mcal{U}_1\} \nn
&=\{i \mid i \in \mcal{C} \in \mathbb{F}(\mcal{U}_2),\> i \not\in \mcal{U}_2\} 
\cup \{i \mid i \in \mcal{C} \in \mathbb{F}(\mcal{U}_2),\> i \in \mcal{Y}\} \nn
&= \partial \mcal{U}_2 \cup \mcal{Y}.
\label{eqn:relation_U1_U2}
\end{align}
The relation
\begin{align}
\rho_{\mcal{T}}(\mcal{U}_2;\bm{x}_{\partial \mcal{U}_2})
&=\sum_{\bm{x}_{\mcal{U}_1}}\sum_{\bm{x}_{\mcal{Y}}}f(\bm{x}_{\mcal{T}}) P\big(\bm{x}_{\mcal{U}_1}, \bm{x}_{\mcal{Y}} \mid \bm{x}_{\partial\mcal{U}_2} \big)\nn
&=\sum_{\bm{x}_{\mcal{U}_1}} \sum_{\bm{x}_{\mcal{Y}}}f(\bm{x}_{\mcal{T}}) P\big(\bm{x}_{\mcal{U}_1}\mid \bm{x}_{\mcal{Y}} ,\bm{x}_{\partial\mcal{U}_2} \big)
P\big(\bm{x}_{\mcal{Y}} \mid \bm{x}_{\partial\mcal{U}_2} \big)\nn
&=\sum_{\bm{x}_{\mcal{Y}}}\rho_{\mcal{T}}(\mcal{U}_1;\bm{x}_{\partial \mcal{U}_1})P\big(\bm{x}_{\mcal{Y}} \mid \bm{x}_{\partial\mcal{U}_2} \big)
\label{eqn:relation_rho}
\end{align}
always holds because
\begin{align*}
P\big(\bm{x}_{\mcal{U}_1}\mid \bm{x}_{\mcal{Y}} ,\bm{x}_{\partial\mcal{U}_2} \big)=P\big(\bm{x}_{\mcal{U}_1}\mid \bm{x}_{\partial \mcal{U}_1}  \big)
\end{align*} 
is satisfied owing to both the spatial Markov property of the HMRF and the relation in equation (\ref{eqn:relation_U1_U2}). 
From equation (\ref{eqn:relation_rho}), the difference in equation (\ref{eqn:diff_rho}) can be rewritten as
\begin{align}
E = \sum_{\bm{x}_{\mcal{Y}}}\sum_{\bm{x}_{\partial \mcal{U}_2}}\big( \rho_{\mcal{T}}(\mcal{U}_1;\bm{x}_{\partial \mcal{U}_1})
-\rho_{\mcal{T}}(\mcal{U}_2;\bm{x}_{\partial \mcal{U}_2})\big)^2 P\big(\bm{x}_{\mcal{Y}} , \bm{x}_{\partial\mcal{U}_2} \big).
\end{align}
This equation indicates that $E$ is always greater than or equal to zero. 
Therefore, it is proved that $v_{\mcal{T}}(\mcal{U}_1) \geq  v_{\mcal{T}}(\mcal{U}_2)\geq 0$.

\section{Derivations of SMCI Methods on PBMs}

\subsection{Derivations of equations (\ref{eqn:e1_1-SMCI}) and (\ref{eqn:e2_1-SMCI})}
\label{app:Derivation_1SMCI}

In general, the 1-SMCI method is represented by 
\begin{align}
m_{\mcal{T}}^{(1)}(\mathbb{S})
= \frac{1}{M}\sum_{\ell=1}^M \sum_{\bm{x}_{\mcal{T}}}f(\bm{x}_{\mcal{T}}) P\big(\bm{x}_{\mcal{T}} \mid \mbf{s}_{\mcal{N}_{1}(\mcal{T})}^{(\ell)} \big).
\label{eqn:1-SMCI}
\end{align}
In the PBM in equation (\ref{eqn:pairwiseMRF}), the conditional distribution $P(\bm{x}_{\mcal{T}} \mid \bm{x}_{\mcal{N}_1(i,j)})$ is represented as
\begin{align}
P(x_i \mid \bm{x}_{\mcal{N}_1(i)}) &= \frac{\exp(\gamma_i(\bm{x}_{\mcal{N}_1(i)}) x_i)}{z_i(\gamma_i(\bm{x}_{\mcal{N}_1(i)}))}, 
\label{eqn:cond_P1(xi)}
\end{align}
when $\mcal{T} = \{i\}$, 
and
\begin{align}
P(x_i, x_j \mid \bm{x}_{\mcal{N}_1(i,j)}) = \frac{\exp (\gamma_{i : j}(\bm{x}_{\mcal{N}_1(i,j)}) x_i + \gamma_{j : i}(\bm{x}_{\mcal{N}_1(i,j)}) x_j  + w_{i,j}x_ix_j)}
{z_{i,j}(\gamma_{i : j}(\bm{x}_{\mcal{N}_1(i,j)}), \gamma_{j : i}(\bm{x}_{\mcal{N}_1(i,j)}), w_{i,j})},
\label{eqn:cond_P1(xi,xj)}
\end{align}
when $\mcal{T} = \{i, j\}$, where 
\begin{align*}
z_i(a) := \sum_{x_i \in \mcal{X}_i} \exp (a x_i),\quad
z_{i,j}(a,b,c):=\sum_{x_i \in \mcal{X}_i}\sum_{x_j \in \mcal{X}_j} \exp (a x_i + bx_j + cx_ix_j).
\end{align*}
$\gamma_i(\bm{x}_{\mcal{N}_1(i)})$ and $\gamma_{i : j}(\bm{x}_{\mcal{N}_1(i,j)})$ in equations (\ref{eqn:cond_P1(xi)}) and (\ref{eqn:cond_P1(xi,xj)}) are defined in equations (\ref{eqn:gamma_i}) and (\ref{eqn:gamma_i-j}). 
Note that $\gamma_{i \setminus j}(\bm{x}_{\partial i})$ does not depend on $x_i$ and $x_j$.

When $\mcal{X}_i = \{-1,+1\}$, the equations
\begin{align}
\sum_{x_i \in \mcal{X}_i} x_i \frac{\exp(a x_i)}{z_i(a)} = \tanh a
\label{eqn:sum-operation_1}
\end{align}
and
\begin{align}
\sum_{x_i \in \mcal{X}_i}\sum_{x_j \in \mcal{X}_j}x_i x_j \frac{\exp (a x_i + bx_j + cx_ix_j)}{z_{i,j}(a,b,c)}
= \tanh \big[ \atanh \big\{ (\tanh a) (\tanh b)\big\} + c \big]
\label{eqn:sum-operation_2}
\end{align}
are obtained. Equations (\ref{eqn:1-SMCI})--(\ref{eqn:sum-operation_2}) lead to the 1-SMCI methods in equations (\ref{eqn:e1_1-SMCI}) and (\ref{eqn:e2_1-SMCI}).

\subsection{Derivations of equations (\ref{eqn:e1_s2-SMCI}) and (\ref{eqn:e2_s2-SMCI})}
\label{app:Derivation_s2SMCI}

When $\mcal{T}=\{i\}$, equation (\ref{eqn:marginal_conditional-dist_GSMCI_PBM}) becomes
\begin{align}
P(\bm{x}_{\mcal{T}} \mid \bm{x}_{\partial \mcal{A}}) \propto
\exp \big( \xi_i(\bm{x}_{\partial \mcal{A}}) x_i \big),
\label{eqn:cond_Ps2(xi)}
\end{align}
where $\xi_i(\bm{x}_{\partial \mcal{A}})$ is defined in equation (\ref{eqn:xi_i}). 
Equation (\ref{eqn:cond_Ps2(xi)}) is obtained by using the equation
\begin{align*}
\ln \cosh (a + b x)= x \atanh\big\{ (\tanh a) (\tanh b)\big\} + \mrm{constant},
\end{align*}
which is satisfied when $x \in \{-1, +1\}$\footnote{
For $x \in \{-1,+1\}$, a function $f(x)$ is always represented as $f(x) = A + Bx$, where $A = \sum_{x \in \{-1,+1\}}  f(x)/2$ and 
$B = \sum_{x \in \{-1,+1\}} x f(x)/2$.}. By using equations (\ref{eqn:GSMCI_s2}), (\ref{eqn:sum-operation_1}), and (\ref{eqn:cond_Ps2(xi)}), 
equation (\ref{eqn:e1_s2-SMCI}) is obtained.

When $\mcal{T}=\{i, j\}$, equation (\ref{eqn:marginal_conditional-dist_GSMCI_PBM}) becomes
\begin{align}
P(\bm{x}_{\mcal{T}} \mid \bm{x}_{\partial \mcal{A}}) &\propto
\exp \big( \xi_{i:j}(\bm{x}_{\partial \mcal{A}}) x_i + \xi_{j:i}(\bm{x}_{\partial \mcal{A}}) x_j 
+\omega_{i,j}(\bm{x}_{\partial \mcal{A}}) x_ix_j  \big),
\label{eqn:cond_Ps2(xi,xj)}
\end{align}
where $\xi_{i:j}(\bm{x}_{\partial \mcal{A}})$ and $\omega_{i,j}(\bm{x}_{\partial \mcal{A}})$ are defined in equations (\ref{eqn:xi_i-j}) and (\ref{eqn:pmega_ij}), respectively. 
Equation (\ref{eqn:cond_Ps2(xi,xj)}) is obtained by using the equation\footnote{
For $x_i, x_j \in \{-1,+1\}$, a function $f(x_i, x_j)$ is always represented as $f(x_i, x_j) = A + Bx_i + Cx_j + Dx_ix_j$, 
where $A = \sum_{x_i,x_j \in \{-1,+1\}}f(x_i, x_j)/4$, $B = \sum_{x_i,x_j \in \{-1,+1\}} x_i f(x_i, x_j)/4$, 
$C = \sum_{x_i,x_j \in \{-1,+1\}} x_j f(x_i, x_j)/4$, and $D = \sum_{x_i,x_j \in \{-1,+1\}} x_i x_j f(x_i, x_j)/4$ 
}
\begin{align*}
&\ln \cosh(a +  b x_i + c x_j) \nn
&= x_i x_j\Big\{ \atanh\big\{ (\tanh b) (\tanh c)\big\} + \frac{1}{4} \ln \frac{1 - (\tanh^2 a) (\tanh^2 (b + c))}{1 - (\tanh^2 a)( \tanh^2 (b - c))}\Big\}\nn
\aleq
+x_i\Big\{ \atanh\big\{ (\tanh a) (\tanh b)\big\} + \frac{1}{4} \ln \frac{1 - (\tanh^2 c) (\tanh^2 (a + b))}{1 - (\tanh^2 c)( \tanh^2 (a - b))}\Big\}\nn
\aleq
+x_j\Big\{ \atanh\big\{ (\tanh a) (\tanh c)\big\} + \frac{1}{4} \ln \frac{1 - (\tanh^2 b) (\tanh^2 (a + c))}{1 - (\tanh^2 b)( \tanh^2 (a - c))}\Big\}\nn
\aleq
+\mrm{constant}, 
\end{align*}
which is satisfied when $x_i, x_j \in \{-1, +1\}$. 
By using equations (\ref{eqn:GSMCI_s2}), (\ref{eqn:sum-operation_2}), and (\ref{eqn:cond_Ps2(xi,xj)}), equation (\ref{eqn:e2_s2-SMCI}) is obtained.

\bibliographystyle{unsrt}
\bibliography{citation}

\end{document}